\documentclass[10pt,twocolumn,letterpaper]{article}

\usepackage{cvpr}
\usepackage{times}
\usepackage{epsfig}
\usepackage{graphicx}
\usepackage{amsmath}
\usepackage{amssymb}
\usepackage{subcaption}
\usepackage{bm}
\usepackage{multirow}
\usepackage{booktabs}
\usepackage{enumitem}

\usepackage[pagebackref=true,breaklinks=true,letterpaper=true,colorlinks,bookmarks=false]{hyperref}

\cvprfinalcopy 

\def\cvprPaperID{115} 

\ifcvprfinal\fi
\begin{document}

\title{Say As You Wish: Fine-grained Control of Image Caption Generation with Abstract Scene Graphs}

\author{Shizhe Chen\textsuperscript{1}\thanks{This work was performed when Shizhe Chen was visiting University of Adelaide.}, Qin Jin\textsuperscript{1}\thanks{Qin Jin is the corresponding author.}, Peng Wang\textsuperscript{2}, Qi Wu\textsuperscript{3}\\
\textsuperscript{1}Renmin University of China, \textsuperscript{2}Northwestern Polytechnical University\\ 
\textsuperscript{3}Australian Centre for Robotic Vision, University of Adelaide\\
{\tt\small \{cszhe1, qjin\}@ruc.edu.cn, peng.wang@nwpu.edu.cn, qi.wu01@adelaide.edu.au}
}

\maketitle

\begin{abstract}
Humans are able to describe image contents with coarse to fine details as they wish.
However, most image captioning models are intention-agnostic which can not generate diverse descriptions according to different user intentions initiatively. 
In this work, we propose the Abstract Scene Graph (ASG) structure to represent user intention in fine-grained level and control what and how detailed the generated description should be.
The ASG is a directed graph consisting of three types of \textbf{abstract nodes} (object, attribute, relationship) grounded in the image without any concrete semantic labels. Thus it is easy to obtain either manually or automatically.
From the ASG, we propose a novel ASG2Caption model, which is able to recognise user intentions and semantics in the graph, and therefore generate desired captions according to the graph structure. 
Our model achieves better controllability conditioning on ASGs than carefully designed baselines on both VisualGenome and MSCOCO datasets. 
It also significantly improves the caption diversity via automatically sampling diverse ASGs as control signals.
\end{abstract}

\section{Introduction}
Image captioning is a complex problem since it requires a machine to complete several computer vision tasks, such as object recognition, scene classification, attributes and relationship detection, simultaneously, and then summarise them to a sentence.
Thanks to the rapid development of deep learning \cite{he2016deep,hochreiter1997long}, recent image captioning models \cite{anderson2018bottom,rennie2017self,xu2015show} have made substantial progress and even outperform humans in terms of several accuracy-based evaluation metrics \cite{banerjee2005meteor,papineni2002bleu,vedantam2015cider}.

\begin{figure}
	\includegraphics[width=1\linewidth]{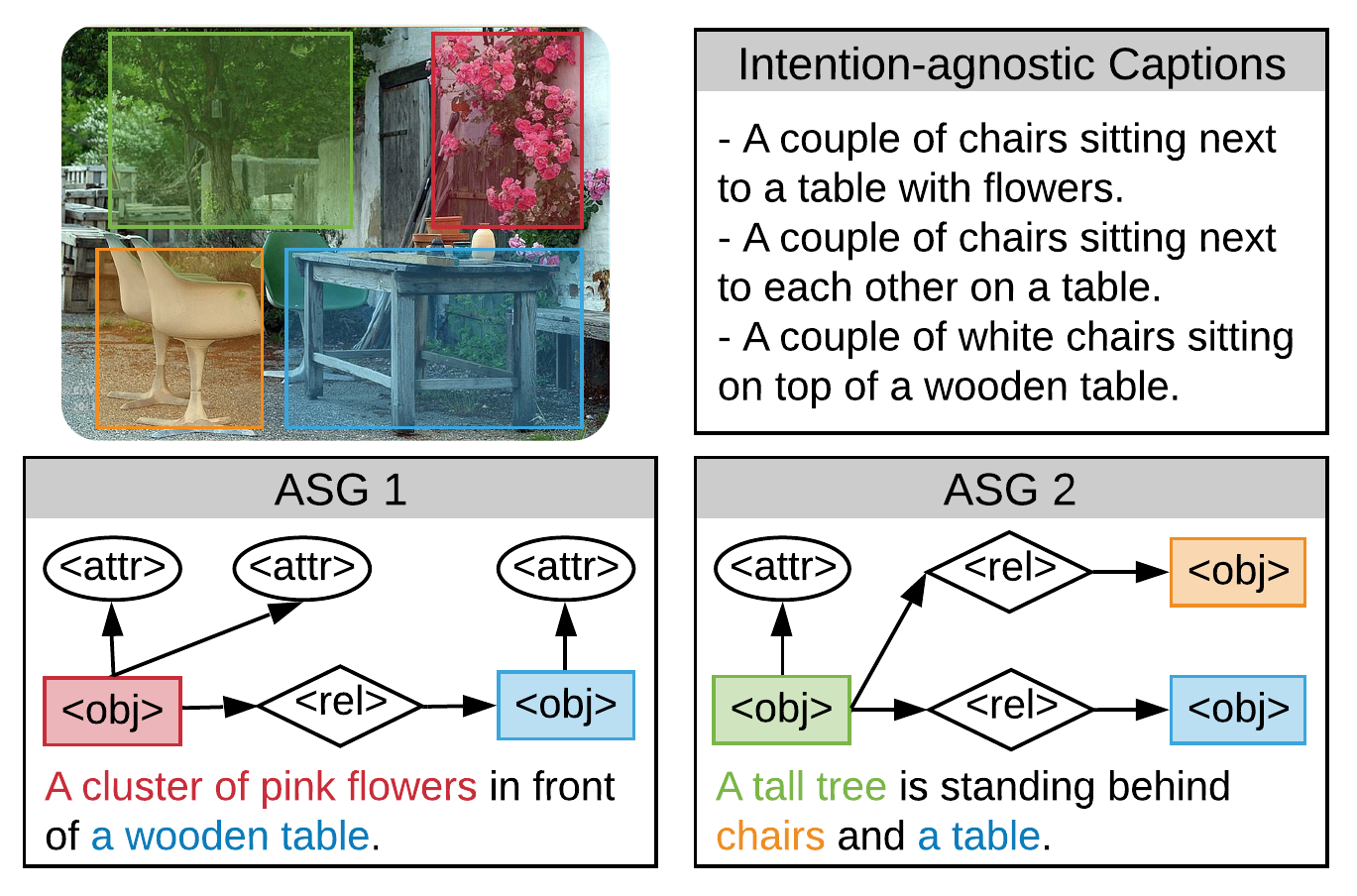}
	\caption{Although intention-agnostic captions can correctly describe image contents, they fail to realise what a user wants to describe and lack of diversity. Therefore, we propose Abstract Scene Graphs (ASG) to control the generation of user desired and diverse image captions in fine-grained level. The corresponding region, ASG node and generated phrase are labelled as the same colour. }
	\label{fig:intro_task}
	\vspace{-10pt}
\end{figure}

However, most image captioning models are intention-agnostic and only passively generate image descriptions, which do not care about what contents users are interested in, and how detailed the description should be.
On the contrary, we humans are able to describe image contents from coarse to fine details as we wish.
For example, we can describe more discriminative details (such as the quantity and colour) of flowers in Figure~\ref{fig:intro_task} if we are asked to do so, but current systems totally fail to realise such user intention.
What is worse, such passive caption generation can greatly hinder diversity and tend to generate mediocre descriptions \cite{shetty2017speaking,wang2019describing}.
Despite achieving high accuracy, these descriptions mainly capture frequent descriptive patterns and cannot represent holistic image understanding, which is supposed to recognise different aspects in the image and thus be able to produce more diverse descriptions.

In order to address aforementioned limitations, few previous endeavours have proposed to actively control image captioning process.
One type of works \cite{gan2017stylenet,guo2019mscap,mathews2018semstyle} focuses on controlling expressive styles of image descriptions such as factual, romantic, humorous styles \etc,
while the other type aims to control the description contents such as different image regions \cite{johnson2016densecap}, objects \cite{Cornia_2019_CVPR,Zheng_2019_CVPR}, and part-of-speech tags \cite{Deshpande_2019_CVPR}, so that the model is able to describe user interested contents in the image.
However, all of the above works can only handle a coarse-grained control signal such as one-hot labels or a set of image regions, which are hard to realise user desired control at a fine-grained level, for instance describing various objects in different level of details as well as their relationships.

In this work, we propose a more fine-grained control signal, \emph{Abstract Scene Graph} (ASG), to represent different intentions for controllable image caption generation.
As shown in Figure~\ref{fig:intro_task}, the ASG is a directed graph consisting of three types of abstract nodes grounded in the image, namely object, attribute and relationship, while no concrete semantic label is necessary for each node. Therefore, such graph structure is easy to obtain either manually or automatically since it does not require semantic recognition.
More importantly, the ASG is capable of reflecting user's fine-grained intention on what to describe and how detailed to describe.

In order to generate captions respecting designated ASGs, we then propose an ASG2Caption model based on an encoder-decoder framework which tackles three main challenges in ASG controlled caption generation.
Firstly, notice that our ASG only contains an abstract scene layout without any semantic labels, it is necessary to capture both intentions and semantics in the graph.
Therefore, we propose a role-aware graph encoder to differentiate fine-grained intention roles of nodes and enhance each node with graph contexts to improve semantic representation.
Secondly, the ASG not only controls what contents to describe via different nodes, but also implicitly decides the descriptive order via how nodes are connected.
Our proposed decoder thus considers both content and structure of nodes for attention to generate desired content in graph flow order.
Last but not least, it is important to fully cover information in ASG without missing or repetition.
For this purpose, our model gradually updates the graph representation during decoding to keep tracking of graph access status.

Since there are no available datasets with ASG annotation, we automatically construct ASGs for training and evaluation on two widely used image captioning datasets, VisualGenome and MSCOCO.
Extensive experiments demonstrate that our approach can achieve better controllability given designated ASGs than carefully designed baselines.
Furthermore, our model is capable of generating more diverse captions based on automatically sampled ASGs to describe various aspects in the image. 

The contributions of our work are three-fold:
\parskip=0.1em
\begin{itemize}[itemsep=0.1em,parsep=0em,topsep=0em,partopsep=0em]
	\item To the best of our knowledge, we are the first to propose fine-grained control of image caption generation with Abstract Scene Graph, which is able to control the level of details (such as, whether attributes, relationships between objects should be included) in the caption generation process.
	\item The proposed ASG2Caption model consists of a \textit{role-aware graph encoder} and \textit{language decoder for graphs} to automatically recognise abstract graph nodes and generate captions with intended contents and orders. 
	\item We achieve state-of-the-art controllability given designated ASGs on two datasets. Our approach can also be easily extended to automatically generated ASGs, which is able to generate diverse image descriptions.
\end{itemize}
\section{Related Work}
\subsection{Image Captioning}
Image captioning \cite{anderson2018bottom,gan2017semantic,vinyals2015show,wu2016value,xu2015show} has achieved significant improvements based on neural encoder-decoder framework \cite{sutskever2014sequence}.
The Show-Tell model \cite{vinyals2015show} employs convolutional neural networks (CNNs) \cite{he2016deep} to encode image into fixed-length vector, and recurrent neural networks (RNNs) \cite{hochreiter1997long} as decoder to sequentially generate words.
To capture fine-grained visual details, attentive image captioning models \cite{anderson2018bottom,lu2017knowing,xu2015show} are proposed to dynamically ground words with relevant image parts in generation.
To reduce exposure bias and metric mismatching in sequential training \cite{ranzato2015sequence}, notable efforts are made to optimise non-differentiable metrics using reinforcement learning \cite{liu2017improved,rennie2017self}.
To further boost accuracy, detected semantic concepts \cite{gan2017semantic,wu2016value,you2016image} are adopted in captioning framework.
The visual concepts learned from large-scale external datasets also enable the model to generate captions with novel objects beyond paired image captioning datasets \cite{agrawal2018nocaps,lu2018neural}.
A more structured representation over concepts, scene graph \cite{johnson2015image}, is further explored \cite{yang2019auto,yao2018exploring} in image captioning which can take advantage of detected objects and their relationships.
In this work, instead of using a fully detected scene graph (which is already a challenging enough task \cite{zellers2018neural,zhang2019vrd}) to improve captioning accuracy, we propose to employ abstract scene graph (ASG) as a control signal to generate desired and diverse image captions.
The ASG is convenient to interact with human to control captioning in fine-grained level, and easier to be obtained automatically than fully detected scene graphs.

\begin{figure*}
	\centering
	\includegraphics[width=1\linewidth]{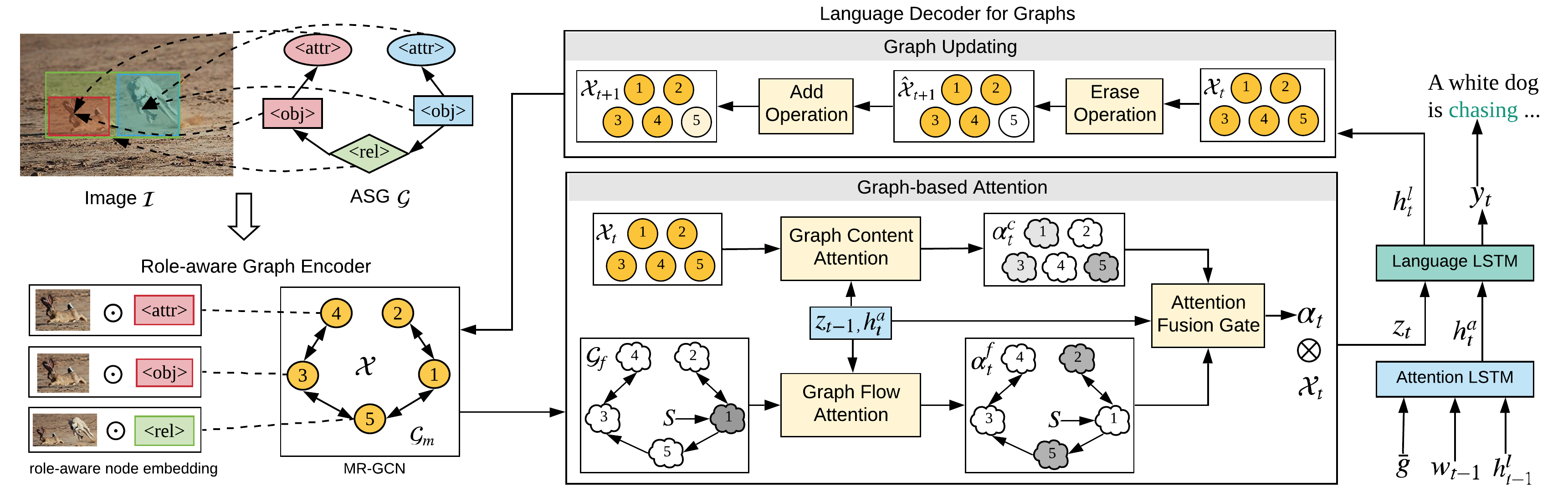}
	\caption{The proposed ASG2Caption model consists of a role-aware graph encoder and a language decoder for graphs. Given an image $\mathcal{I}$ and ASG $\mathcal{G}$, our encoder first initialises each node as role-aware embedding, and employs a multi-layer MR-GCN to encode graph contexts in $\mathcal{G}_m$. Then the decoder dynamically incorporates graph content and graph flow attentions for ASG-controlled captioning. After generating a word, we update the graph $\mathcal{X}_{t-1}$ into $\mathcal{X}_{t}$ to record graph access status.}
	\label{fig:method_framework}
	\vspace{-10pt}
\end{figure*}

\subsection{Controllable Image Caption Generation}
Controllable text generation \cite{hu2017toward,keskarCTRL2019} aims to generate sentences conditioning on designated control signals such as sentiment, styles, semantic \etc, which is more interactive, interpretable and easier to generate diverse sentences.
There are broadly two groups of control for image captioning, namely style control and content control. 
The style control researches \cite{gan2017stylenet,guo2019mscap,mathews2018semstyle,mathews2016senticap} aim to describe global image content with different styles.
The main challenge is lack of paired stylised texts for training.
Therefore, recent works \cite{gan2017stylenet,guo2019mscap,mathews2018semstyle} mainly disentangle style codes from semantic contents so that unpaired style transfer can be applied. 

The content control works  \cite{Cornia_2019_CVPR,johnson2016densecap,yang2017dense,Zheng_2019_CVPR} instead aim to generate captions capturing different aspects in the image such as different regions, objects and so on, which are more relevant to holistic visual understanding.
Johnson \etal \cite{johnson2016densecap} is the first to propose the dense captioning task, which detects and describes diverse regions in the image.
Zheng \etal \cite{Zheng_2019_CVPR} constrain the model to involve a human concerned object.
Cornia \etal \cite{Cornia_2019_CVPR} further control multiple objects and their orders in the generated description.
Besides manipulating on object-level, Deshpande \etal \cite{Deshpande_2019_CVPR} employ Part-of-Speech (POS) syntax to guide caption generation, which however mainly focus on improving diversity rather than POS control.
Beyond single image, Park \etal \cite{Park_2019_ICCV} propose to only describe semantic differences between two images.

However, none of above works can control caption generation at more fine-grained level. For instance, whether (and how many) associative attributes should be used? Any other objects (and its associated relationships) should be included and what is the description order? 
In this paper, we propose to utilise fine-grained ASG to control designated structure of objects, attributes and relationships at the same time, and enable generating more diverse captions that reflect different intentions.

\section{Abstract Scene Graph}
\label{sec:method_sgl}
In order to represent user intentions in fine-grained level, we first propose an Abstract Scene Graph (ASG) as the control signal for generating customised image captions.
An ASG for image $\mathcal{I}$ is denoted as $\mathcal{G=(V, E)}$, where $\mathcal{V}$ and $\mathcal{E}$ are the sets of nodes and edges respectively. 
As illustrated in the top left of Figure~\ref{fig:method_framework}, the nodes can be classified into three types according to their intention roles: object node $o$, attribute node $a$ and relationship node $r$.
The user intention is constructed into $\mathcal{G}$ as follows:
\parskip=0.1em
\begin{itemize}[itemsep=0.1em,parsep=0em,topsep=0em,partopsep=0em]
	\item add user interested object $o_i$ to $\mathcal{G}$, where object $o_i$ is grounded in $\mathcal{I}$ with a corresponding bounding box;
	\item if the user wants to know more descriptive details of $o_i$,  add an attribute node $a_{i, l}$ to $\mathcal{G}$ and assign a directed edge from $o_i$ to $a_{i, l}$. $|l|$ is the number of associative attributes since multiple $a_{i, l}$ for $o_i$ are allowed;
	\item if the user wants to describe relationship between $o_i$ and $o_j$, where $o_i$ is the subject and $o_j$ is the object, add relationship node $r_{i, j}$ to $\mathcal{G}$ and assign  directed edges from $o_i$ to $r_{i, j}$ and from $r_{i, j}$ to $o_j$ respectively.
\end{itemize}
It is convenient for an user to construct the ASG $\mathcal{G}$, which represents the user's interests about objects, attributes and relationships in $\mathcal{I}$ in a fine-grained manner. 

Besides obtaining $\mathcal{G}$ from users, it is also easier to generate ASGs automatically based on off-the-shelf object proposal networks and optionally a simple relationship classifier to tell whether two objects contain any relationship. Notice that our ASG is only a graph layout without any semantic labels, which means we do not rely on externally trained object/attribute/relationship detectors, but previous scene graph based image captioning models \cite{yao2018exploring} need these well-trained detectors to provide a complete scene graph with labels and have to suffer the low detection accuracy.

The details of automatic ASG generation are provided in the supplementary material.
In this way, diverse ASGs can be extracted to capture different aspects in the image and thus lead to diverse caption generation.

\section{The ASG2Caption Model}
Given an image $\mathcal{I}$ and a designated ASG $\mathcal{G}$, the goal is to generate a fluent sentence $y=\{y_1, \cdots, y_T\}$ that strictly aligns with $\mathcal{G}$ to satisfy user's intention.
In this section, we present the proposed ASG2Caption model which is illustrated in Figure~\ref{fig:method_framework}. 
We will describe the proposed encoder and decoder in Section~\ref{sec:method_encoder} and \ref{sec:method_decoder} respectively, followed by its training and inference strategies in Section~\ref{method_train}.

\subsection{Role-aware Graph Encoder}
\label{sec:method_encoder}
The encoder is proposed to encode ASG $\mathcal{G}$ grounded in image $\mathcal{I}$ as a set of node embeddings $\mathcal{X}=\{x_1, \cdots, x_{|\mathcal{V}|}\}$.
Firstly, $x_i$ is supposed to reflect its intention role besides the visual appearance, which is especially important to differentiate object and connected attribute nodes since they are grounded in the same region.
Secondly, since nodes are not isolated, contextual information from neighboured nodes is beneficial to recognise semantic meaning of the node.
Therefore, we propose a role-aware graph encoder, which contains a \textit{role-aware node embedding} to distinguish node intentions and a \textit{multi-relational graph convolutional network (MR-GCN)} ~\cite{schlichtkrull2018modeling} for contextual encoding.

\paragraph{Role-aware Node Embedding.}
For the $i$-th node in $\mathcal{G}$, we firstly initialise it as a corresponding visual feature $v_i$.
Specifically, the feature of object node is extracted from the grounded bounding box in the image; the feature of attribute node is the same as its connected object; and the feature of relationship node is extracted from the union bounding box of the two involved objects.
Since visual features alone cannot distinguish intention roles of different nodes, we further enhance each node with role embedding to obtain a role-aware node embedding $x^{(0)}_i$ as follows:
\begin{equation}
	x^{(0)}_i =\left\{
	\begin{array}{rcl}
	& v_i \odot W_r[0],      & {\text{if } i \in o ;}\\
	& v_i \odot (W_r[1] + \mathrm{pos}[i]),     & {\text{if } i \in a;} \\
	& v_i \odot W_r[2],    & {\text{if } i \in r.}
	\end{array} \right.
\end{equation}
where $W_r \in \mathbb{R}^{3 \times d}$ is the role embedding matrix, $d$ is the feature dimension,
$W_r[k]$ denotes the $k$-th row of $W_r$,
and $\mathrm{pos}[i]$ is a positional embedding to distinguish the order of different attribute nodes connected with the same object.

\paragraph{Multi-relational Graph Convolutional Network.}
Though the edge in ASG is uni-directional, the influence between connected nodes is mutual.
Furthermore, since nodes are of different types, how the message passing from one type of node to another is  different from its inverse direction.
Therefore, we extend the original ASG with different bidirectional edges, which leads to a multi-relational graph $\mathcal{G}_m=\{\mathcal{V, E, R} \}$ for contextual encoding.

Specifically, there are six types of edges in  $\mathcal{R}$ to capture mutual relations between neighboured nodes, which are: object to attribute, subject to relationship, relationship to object and their inverse directions respectively.
We employ a MR-GCN to encode graph context in  $\mathcal{G}_m$ as follows:
\begin{equation}
	x^{(l+1)}_i = \sigma (W^{(l)}_0 x^{(l)}_i + \sum_{\tilde{r} \in \mathcal{R}} \sum_{j \in \mathcal{N}_i^{\tilde{r}}} \frac{1}{|\mathcal{N}_i^{\tilde{r}}|} W_{\tilde{r}}^{(l)} x^{(l)}_j)
\end{equation}
where $\mathcal{N}_i^{\tilde{r}}$ denotes neighbours of $i$-th node under relation $\tilde{r} \in \mathcal{R}$, $\sigma$ is the ReLU activation function, and $W^{(l)}_{*}$ are parameters to be learned at $l$-th MR-GCN layer.
Utilising one layer brings contexts from direct neighboured nodes for each node, while stacking multiple layers enables to encode broader contexts in the graph.
We stack $L$ layers and then the outputs of the final $L$-th layer are employed as our final node embeddings $\mathcal{X}$.
We can also obtain a global graph embedding via taking an average of $\mathcal{X}$ as $\bar{g}=\frac{1}{|\mathcal{V}|} \sum_{i} x_i$. 
We fuse global graph embedding with global image representation as global encoded feature $\bar{v}$.

\subsection{Language Decoder for Graphs}
\label{sec:method_decoder}
The decoder aims to convert the encoded $\mathcal{G}$ into an image caption.
Unlike previous works that attend on a set of unrelated vectors \cite{lu2017knowing,xu2015show}, our node embeddings $\mathcal{X}$ contain structured connections from $\mathcal{G}$, which reflects user designated order that should not be ignored. 
Furthermore, in order to fully satisfy user intention, it is important to express all the nodes in $\mathcal{G}$ without missing or repetition, while previous attention methods \cite{lu2017knowing,xu2015show} hardly consider accessed status of attended vectors.
Therefore, in order to improve the graph to sentence quality, we propose a language decoder specifically for graphs, which includes a \textit{graph-based attention mechanism} that considers both graph semantics and structures, and a \textit{graph updating mechanism} that keeps a record of what has been described or not.

\paragraph{Overview of the Decoder.}
The decoder employs a two-layer LSTM structure \cite{anderson2018bottom}, including an attention LSTM and a language LSTM.
The attention LSTM takes the global encoded embedding $\bar{v}$, previous word embedding $w_{t-1}$ and previous output from language LSTM $h^l_{t-1}$ as input to compute an attentive query $h^a_t$:
\begin{equation}
h^a_t = \mathrm{LSTM}([\bar{v}; w_{t-1}; h^l_{t-1}], h^a_{t-1}; \theta^a)
\end{equation}
where $[;]$ is vector concatenation and $\theta^a$ are parameters.

We denote node embeddings at $t$-th step as $\mathcal{X}_t=\{x_{t,1}, \cdots, x_{t,|\mathcal{V}|}\}$ where $\mathcal{X}_1$ is the output of encoder $\mathcal{X}$. The $h^a_t$ is used to retrieve a context vector $z_t$ from $\mathcal{X}_t$ via the proposed graph-based attention mechanism.
Then language LSTM is fed with $z_t$ and $h^a_t$ to generate word sequentially:
\begin{align}
h^l_t &= \mathrm{LSTM}([z_t; h^a_{t}], h^l_{t-1}; \theta^l) \\
p(y_t|y_{<t}) &= \mathrm{softmax}(W_p h^l_t + b_p)
\end{align}
where $\theta^l, W_p, b_p$ are parameters.
After generating word $y_t$, we update node embeddings $\mathcal{X}_t$ into $\mathcal{X}_{t+1}$ via the proposed graph updating mechanism to record new graph access status.
We will explain the graph-based attention and graph updating mechanisms in details in the following sections.

\begin{figure}
	\centering
	\includegraphics[width=1\linewidth]{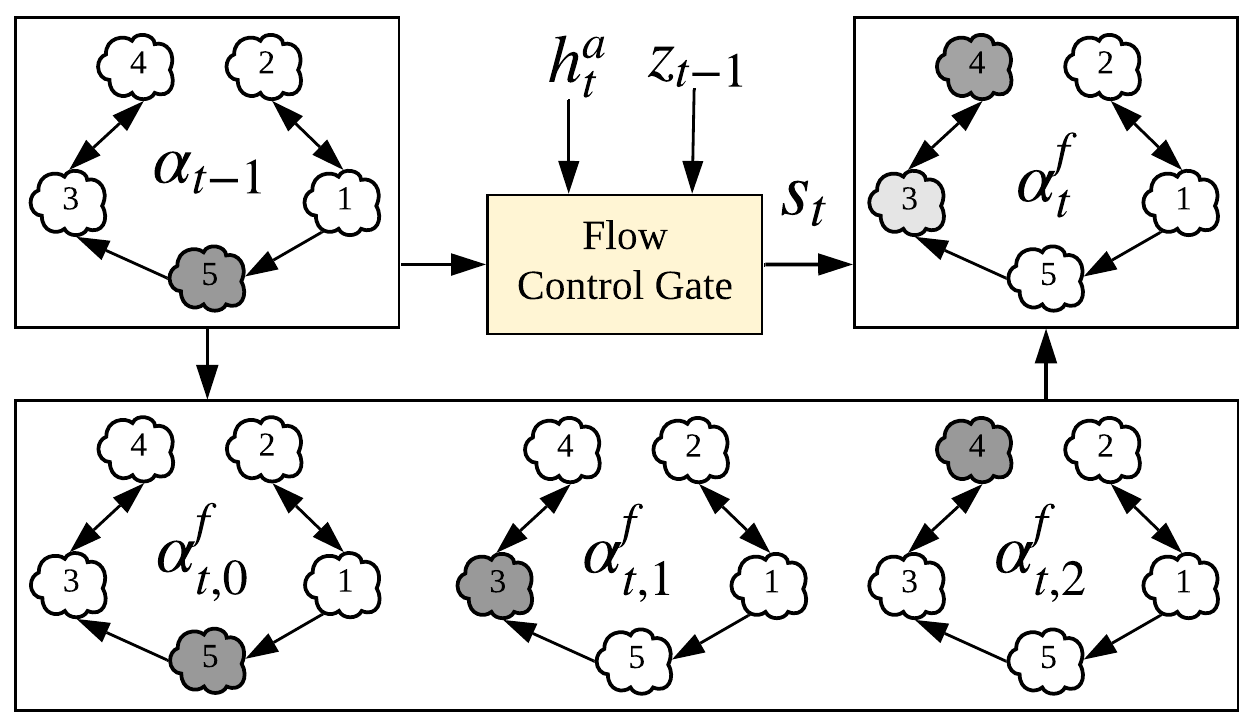}
	\caption{Graph flow attention employs graph flow order to select relevant nodes to generate next word.}
	\label{fig:graph_flow_attn}
\end{figure}

\paragraph{Graph-based Attention Mechanism.}
In order to take into account both semantic content and graph structure, we combine two types of attentions called \emph{graph content attention} and \emph{graph flow attention} respectively.

The graph content attention considers semantic relevancy between node embeddings $\mathcal{X}_t$ and the query $h^a_t$ to compute an attention score vector $\bm{\alpha^c_{t}}$, which is:
\begin{align}
	\tilde{\alpha}^c_{t,i} &= w_c^{T} \mathrm{tanh} (W_{xc} x_{t,i} + W_{hc} h^a_t) \\
	\bm{\alpha^c_{t}} &= \mathrm{softmax} (\bm{\tilde{\alpha}^c_t})
\end{align}
where $W_{xc}, W_{hc}, w_c$ are parameters in content attention and we omit the bias term for simplicity.
Since connections between nodes are ignored, the content attention is similar to teleport which can transfer from one node to another node in far distance in $\mathcal{G}$ at different decoding timesteps.

However, the structure of ASG implicitly reflects the user intended orders on caption generation.
For example, if the current attended node is a relationship node, then the next node to be accessed is most likely to be the following object node according to the graph flow.
Therefore, we further propose a graph flow attention to capture the graph structure.
The flow graph $\mathcal{G}_f$ is illustrated in Figure~\ref{fig:method_framework}, which is different from the original ASG in three ways.
The first is that a start symbol $S$ should be assigned and the second difference lies in the bidirectional connection between object node and attribute node since in general the order of objects and their attributes are not compulsive and should be decided by sentence fluency.
Finally, a self-loop edge will be constructed for a node if there exists no output edge of the node, which ensures the attention on the graph doesn't vanish.
Suppose $M_f$ is the adjacent matrix of the flow graph $\mathcal{G}_f$, where the $i$-th row denotes the normalised in-degree of the $i$-th node.
The graph flow attention transfers attention score vector in previous decoding step $\bm{\alpha_{t-1}}$ in three ways:
\begin{enumerate}
	\item [1)] stay at the same node $\bm{\alpha^f_{t, 0}} = \bm{\alpha_{t-1}}$. For example, the model might express one node with multiple words;
	\item [2)] move one step $\bm{\alpha^f_{t, 1}} = M_f \bm{\alpha_{t-1}}$, for instance transferring from a relationship node to its object node;
	\item [3)] move two steps $\bm{\alpha^f_{t, 2}} = (M_f)^2 \bm{\alpha_{t-1}}$ such as transferring from a relationship node to an attribute node.
\end{enumerate}
The final flow attention is a soft interpolation of the three flow scores controlled by a dynamic gate as follows:
\begin{align}
	\bm{s_t} &= \mathrm{softmax} (W_s \sigma(W_{sh} h^a_t + W_{sz} z_{t-1}))\\
	\bm{\alpha^f_{t}} &= \sum_{k=0}^{2} s_{t, k} \bm{\alpha^f_{t, k}}
\end{align}
where $W_s, W_{sh}, W_{sz}$ are parameters and $\bm{s_t} \in \mathbb{R}^{3}$.
Figure~\ref{fig:graph_flow_attn} presents the process of graph flow attention.

Our graph-based attention dynamically fuses the graph content attention $\bm{\alpha_t^c}$ and the graph flow attention $\bm{\alpha_t^f}$ with learnable parameters $w_g, W_{gh}, W_{gz}$, which is:
\begin{align}
	\beta_t &= \mathrm{sigmoid} (w_g \sigma(W_{gh} h^a_t + W_{gz} z_{t-1})) \\ 
	\bm{\alpha_t} &= \beta_{t}\bm{\alpha^{c}_{t}} + (1 - \beta_t) \bm{\alpha^{f}_{t}}
\end{align}
Therefore, the context vector for predicting word at the $t$-th step is $z_t = \sum_{i=1}^{|\mathcal{V}|} \alpha_{t, i} x_{t, i}$, which is a weighted sum of graph node features.

\begin{table*}
	\centering
	\small
	\caption{Statistics of VisualGenome and MSCOCO datasets for controllable image captioning with ASGs.}
	\label{tab:dataset_stats}
	\begin{tabular}{c|cc|cc|cc|cccc} \toprule
		\multirow{2}{*}{dataset} & \multicolumn{2}{c|}{train} & \multicolumn{2}{c|}{validation} & \multicolumn{2}{c|}{test} & \multirow{2}{*}{\begin{tabular}[c]{@{}c@{}}\#objs\\ per sent\end{tabular}} & \multirow{2}{*}{\begin{tabular}[c]{@{}c@{}}\#rels\\ per sent\end{tabular}} & \multirow{2}{*}{\begin{tabular}[c]{@{}c@{}}\#attrs\\ per obj\end{tabular}} & \multirow{2}{*}{\begin{tabular}[c]{@{}c@{}}\#words\\ per sent\end{tabular}} \\
		& \#imgs & \#sents & \#imgs & \#sents & \#imgs & \#sents &  &  &  &  \\ \midrule
		VisualGenome & 96,738 & 3,397,459 & 4,925 & 172,290 & 4,941 & 171,759 & 2.09 & 0.95 & 0.47 & 5.30 \\
		MSCOCO & 112,742 & 475,117 & 4,970 & 20,851 & 4,979 & 20,825 & 2.93 & 1.56 & 0.51 & 10.28 \\ \bottomrule
	\end{tabular}
\end{table*}

\begin{table*}
	\centering
	\footnotesize
	\caption{Comparison with carefully designed baselines for controllable image caption generation conditioning on ASGs.}
	\label{tab:control_sota_cmpr}
	\begin{tabular}{l|ccccc|cccc|ccccc|cccc} \toprule
		\multirow{2}{*}{Method} & \multicolumn{9}{c|}{VisualGenome} & \multicolumn{9}{c}{MSCOCO} \\
		 & B4 & M & R & C & S & G & G$_o$ & G$_a$ & G$_r$ & B4 & M & R & C & S & G & G$_o$ & G$_a$ & G$_r$  \\ \midrule
		ST \cite{vinyals2015show} & 11.1 & 17.0 & 34.5 & 139.9 & 31.1 & 1.2 & 0.5 & 0.7 & 0.5 & 10.5 & 16.8 & 36.2 & 100.6 & 24.1 & 1.8 & 0.8 & 1.1 & 1.0 \\
		BUTD \cite{anderson2018bottom} & 10.9 & 16.9 & 34.5 & 139.4 & 31.4 & 1.2 & 0.5 & 0.7 & 0.5 & 11.5 & 17.9 & 37.9 & 111.2 & 26.4 & 1.8 & 0.8 & 1.1 & 1.0 \\ \midrule
		C-ST & 12.8 & 19.0 & 37.6 & 157.6 & 36.6 & 1.1 & 0.4 & 0.7 & 0.4 & 14.4 & 20.1 & 41.4 & 135.6 & 32.9 & 1.6 & 0.6 & 1.0 &  0.8 \\
		C-BUTD & 12.7 & 19.0 & 37.9 & 159.5 & 36.8 & 1.1 & 0.4 & 0.7 & 0.4 & 15.5 & 20.9 & 42.6 & 143.8 & 34.9 & 1.5 & 0.6 & 1.0 & 0.8 \\ \midrule
		Ours & \textbf{17.6} & \textbf{22.1} & \textbf{44.7} & \textbf{202.4} & \textbf{40.6} & \textbf{0.7} & \textbf{0.3} & \textbf{0.3} & \textbf{0.3} & \textbf{23.0} & \textbf{24.5} & \textbf{50.1} & \textbf{204.2} & \textbf{42.1} & \textbf{0.7} & \textbf{0.4} & \textbf{0.3} & \textbf{0.3} \\ \bottomrule
	\end{tabular}
\end{table*}

\vspace{-10pt}
\paragraph{Graph Updating Mechanism.}
We update the graph representation to keep a record of accessed status for different nodes in each decoding step.
The attention score $\bm{\alpha_t}$ indicates accessed intensity of each node so that highly attended node is supposed to be updated more.
However, when generating some non-visual words such as ``the'' and ``of'', though graph nodes are accessed, they are not expressed by the generated word and thus should not be updated.
Therefore, we propose a visual sentinel gate as \cite{lu2017knowing} to adaptively modify the attention intensity as follows:
\begin{equation}
\bm{u_t} = \mathrm{sigmoid}(f_{vs}(h^l_t; \theta_{vs})) \bm{\alpha_t}
\end{equation}
where we implement $f_{vs}$ as a fully connected network parametrised by $\theta_{vs}$ which outputs a scalar to indicate whether attended node is expressed by the generated word.

The updating mechanism for each node is decomposed into two parts: an erase followed by an add operation inspired by NTM \cite{graves2014neural}.
Firstly, the $i$-th graph node representation $x_{t, i}$ is erased according to its update intensity $u_{t, i}$ in a fine-grained way for each feature dimension:
\begin{align}
	e_{t, i} &= \mathrm{sigmoid} (f_{ers}([h^l_t; x_{t, i}]; \theta_{ers}))\\ 
	\hat{x}_{t+1, i} &= x_{t, i} (1 - u_{t, i} e_{t, i})
\end{align}
Therefore, a node can be set as zero if it is no longer need to be accessed.
In case a node might need multiple access and track its status, we also employ an add update operation:
\begin{align}
	a_{t, i} &= \sigma (f_{add}([h^l_t; x_{t, i}]; \theta_{add})) \\
	x_{t+1, i} &= \hat{x}_{t+1, i} + u_{t, i} a_{t, i}
\end{align}
where $f_{ers}$ and $f_{add}$ are fully connected networks with different parameters.
In this way, we update the graph embeddings $\mathcal{X}_t$ into $\mathcal{X}_{t+1}$ for the next decoding step.

\subsection{Training and Inference}
\label{method_train}
We utilise the standard cross entropy loss to train our ASG2Caption model. The loss for a single pair ($\mathcal{I}$, $\mathcal{G}$, $y$) is:
\begin{equation}
	L = - \mathrm{log} \sum_{t=1}^{T} p(y_t|y_{<t}, \mathcal{G}, \mathcal{I})
\end{equation}
After training, our model can generate controllable image captions given the image and designated ASG obtained manually or automatically as described in Section~\ref{sec:method_sgl}.

\section{Experiments}

\subsection{Datasets}
We automatically construct triplets of (image $\mathcal{I}$, ASG $\mathcal{G}$, caption $y$) based on annotations of two widely used image captioning datasets, VisualGenome \cite{krishna2017visual} and MSCOCO \cite{lin2014microsoft}.
Table~\ref{tab:dataset_stats} presents statistics of the two datasets. 

\noindent
\textbf{VisualGenome} contains object annotations and dense regions descriptions.
To obtain ASG for corresponding caption and region, we firstly use a Stanford sentence scene graph parser \cite{schuster2015generating} to parse groundtruth region caption to a scene graph. We then ground objects from the parsed scene graph to object regions according to their locations and semantic labels. After aligning objects, we remove all the semantic labels from the scene graph, and only keep the graph layout and nodes type. More details are in the supplementary material.
We follow the data split setting in \cite{anderson2018bottom}. 

\noindent
\textbf{MSCOCO} dataset contains more than 120,000 images and each image is annotated with around five descriptions. We use the same way as VisualGenome to get ASGs for training.
We adopt the `Karpathy' splits setting \cite{karpathy2015deep}.
As shown in Table~\ref{tab:dataset_stats}, the ASGs in MSCOCO are more complex than those in VisualGenome dataset since they contain more relationships and the captions are longer.

\begin{table*}
	\centering
	\small
	\caption{Ablation study to demonstrate contributions from different proposed components. (role: role-aware node embedding; rgcn: MR-GCN; ctn: graph content attention; flow: graph flow attention; gupdt: graph updating; bs: beam search)}
	\label{tab:control_ablation}
	\begin{tabular}{c|cc|cccc|ccccc|ccccc} \toprule
		& \multicolumn{2}{c|}{Enc} & \multicolumn{4}{c|}{Dec} & \multicolumn{5}{c|}{VisualGenome} & \multicolumn{5}{c}{MSCOCO} \\
		\# & role & rgcn & ctn & flow & gupdt & bs & B4 & M & R & C & S & B4 & M & R & C & S \\ \midrule
		1 &  &  & & &  &  & 11.2 & 18.3 & 36.7 & 146.9 & 35.6 & 13.6 & 19.7 & 41.3 & 130.2 & 32.6 \\
		2 &  &  & \checkmark & &  &  & 10.7 & 18.2 & 36.9 & 146.3 & 35.5 & 14.5 & 20.4 & 42.2 & 135.7 & 34.6 \\ \midrule
		3 & \checkmark &  & \checkmark & &  &  & 14.2 & 20.5 & 40.9 & 176.9 & 38.1 & 18.2 & 22.5 & 44.9 & 166.9 & 37.8 \\
		4 & \checkmark & \checkmark & \checkmark & &  &  & 15.7 & 21.4 & 43.6 & 191.7 & 40.0 & 21.6 & 23.7 & 48.6 & 190.5 & 40.9 \\ \midrule
		5 & \checkmark & \checkmark & \checkmark & \checkmark &  &  & 15.9 & 21.5 & 44.0 & 193.1 & 40.1 & 22.3 & 24.0 & 49.4 & 196.2 & 41.5 \\
		6 & \checkmark & \checkmark & \checkmark &  & \checkmark  &  & 15.8 & 21.4 & 43.5 & 191.6 & 39.9 & 21.8 & 24.1 & 49.1 & 194.2 & 41.4 \\ \midrule
		7 & \checkmark & \checkmark & \checkmark & \checkmark & \checkmark  &  & 16.1 & 21.6 & 44.1 & 194.4 & 40.1 & 22.6 & 24.4 & 50.0 & 199.8 & 41.8 \\
		8 & \checkmark & \checkmark & \checkmark & \checkmark & \checkmark  & \checkmark & \textbf{17.6} & \textbf{22.1} & \textbf{44.7} & \textbf{202.4} & \textbf{40.6} & \textbf{23.0} & \textbf{24.5} & \textbf{50.1} & \textbf{204.2} & \textbf{42.1} \\ \bottomrule
	\end{tabular}
\end{table*}

\subsection{Experimental Settings}
\noindent
\textbf{Evaluation Metrics.}
We evaluate caption qualities in terms of two aspects, \emph{controllability} and \emph{diversity} respectively.
To evaluate the controllability given ASG, we utilise ASG aligned with groundtruth image caption as control signal.
The generated caption is evaluated against groundtruth via five automatic metrics including BLEU \cite{papineni2002bleu}, METEOR \cite{banerjee2005meteor}, ROUGE \cite{lin2004rouge}, CIDEr \cite{vedantam2015cider} and SPICE \cite{anderson2016spice}.
Generally, those scores are higher if semantic recognition is correct and sentence structure aligns better with the ASG.
We also propose a Graph Structure metric $G$ based on SPICE \cite{anderson2016spice} to purely evaluate whether the structure is faithful to ASG.
It measures difference of numbers for $(o), (o, a)$ and $(o, r, o)$ pairs respectively between generated and groundtruch caption, where the lower the better. 
We also break down the overall score $G$ for each type of pairs as $G_o$, $G_a$, $G_r$ respectively.
More details are in the supplementary material.

For the diversity measurement, we first sample the same number of image captions for each model, and evaluate the diversity of sampled captions using two types of metrics:
1) $n$-gram diversity (Div-$n$): a widely used metric \cite{Deshpande_2019_CVPR,aneja2019sequential} which is the ratio of distinct $n$-grams to the total number of words in the best 5 sampled captions;
2) SelfCIDEr \cite{wang2019describing}: a recent metric to evaluate semantic diversity derived from latent semantic analysis and kernelised to use CIDEr similarity.
The higher scores the more diverse captions are.

\noindent
\textbf{Implementation Details.}
We employ Faster-RCNN \cite{ren2015faster} pretrained on VisualGenome to extract visual features for grounded nodes in ASG and ResNet152 pretrained on ImageNet \cite{deng2009imagenet} to extract global image representations.
For role-aware graph encoder, we set the feature dimension as 512 and $L$ as 2.
For language decoder, the word embedding and hidden size of LSTM layers are set to be 512.
During training, the learning rate is 0.0001 with batch size of 128.
In the inference phrase, we utilise beam search with beam size of 5 if not specified.

\subsection{Evaluation on Controllability}
We compare the proposed approach with two groups of carefully designed baselines.
The first group contains traditional intention-agnostic image captioning models, including: 1) Show-Tell (ST) \cite{vinyals2015show} which employs a pretrained Resnet101 as encoder to extract global image representation and an LSTM as decoder; and 2) state-of-the-art BottomUpTopDown (BUTD) model \cite{anderson2018bottom} which dynamically attends over relevant image regions when generating different words.
The second group of models extend the above approaches for ASG-controlled image captioning.
For the non-attentive model (C-ST), we fuse global graph embedding $\bar{g}$ with the original feature; while for the attentive model (C-BUTD), we make the model attend to graph nodes in ASG instead of all detected image regions.

\begin{figure}
    \centering
	\includegraphics[width=1\linewidth]{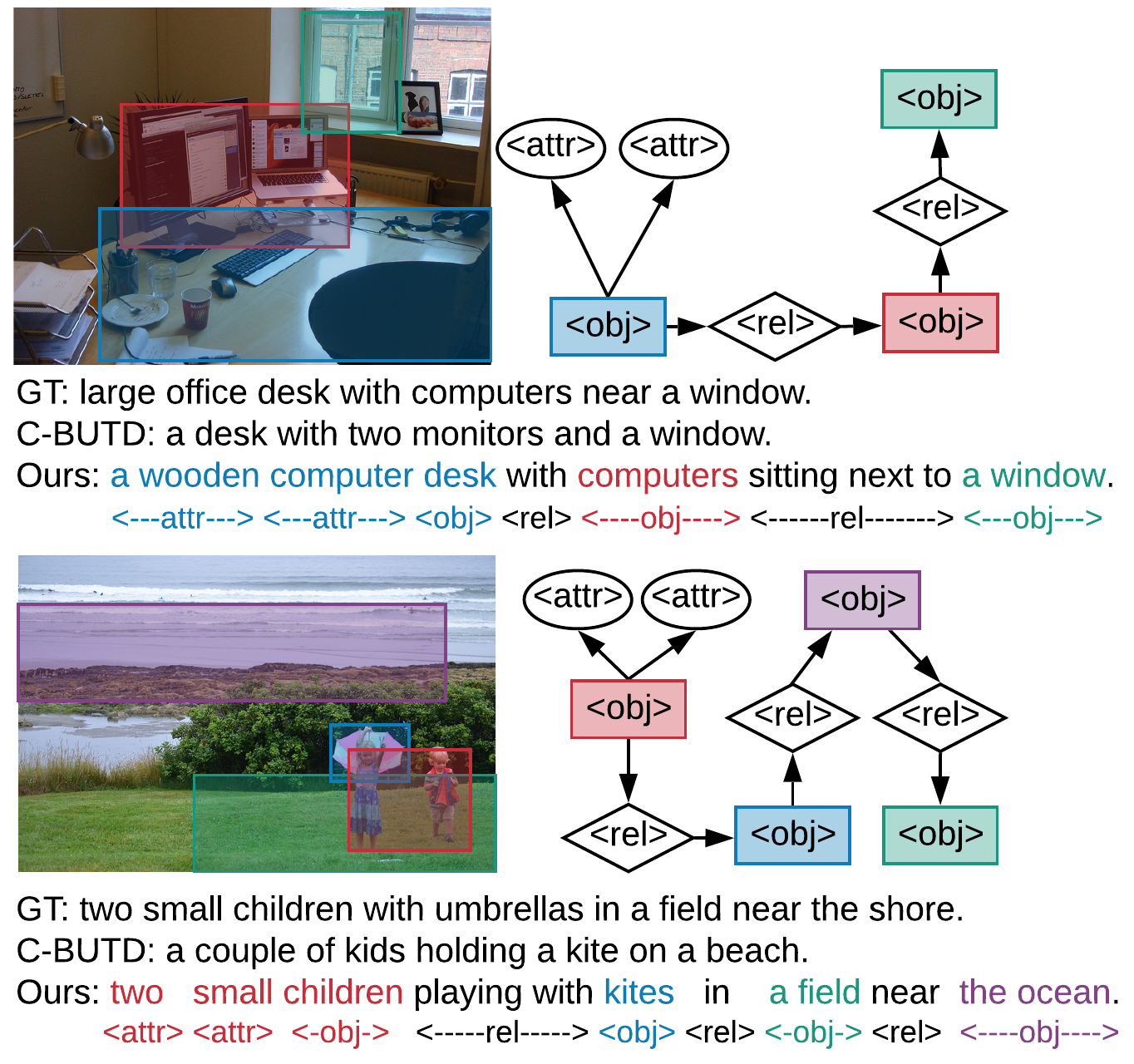}
	\caption{Examples on controllability given designated ASG for different captioning models.}
	\label{fig:control_examples}
	\vspace{-16pt}
\end{figure}

Table~\ref{tab:control_sota_cmpr} presents the comparison result.
It is worth noting that controllable baselines outperform non-controllable baselines due to the awareness of control signal ASG.
We can also see that baseline models are struggling to generate designated attributes compared to objects and relationships according to detailed graph structure metrics.
Our proposed method significantly improves performance than compared approaches on all evaluation metrics in terms of both overall caption quality and alignment with graph structure.
Especially for fine-grained attribute control, we reduce more than half of misalignment on VisualGenome (0.7 $\rightarrow$ 0.3) and MSCOCO (1.0 $\rightarrow$ 0.3) dataset.
In Figure~\ref{fig:control_examples}, we visualise some examples of our ASR2Caption model and the best baseline model C-BUTD.
Our model is more effective to follow designated ASGs for caption generation than C-BUTD model.
In the bottom image of Figure~\ref{fig:control_examples}, though both models fail to recognise the correct concept ``umbrella'', our model still successfully aligns with the graph structure.

\begin{figure*}
    \centering
	\includegraphics[width=1\linewidth]{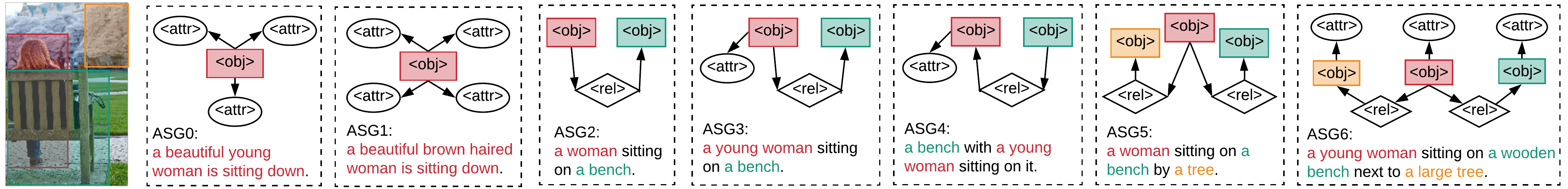}
	\caption{Generated image captions using user created ASGs for the leftmost image. Even subtle changes in the ASG represent different user intentions and lead to different descriptions. Best viewed in colour.}
	\label{fig:user_intention_examples}
\end{figure*}

\begin{figure*}
    \centering
	\includegraphics[width=1\linewidth]{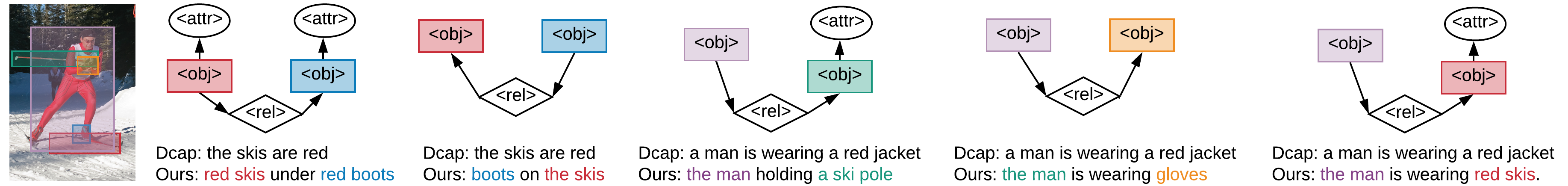}
	\caption{Examples for diverse image caption generation conditioning on sampled ASGs. Our generated captions are different from each other while the comparison baseline (dense-cap) generates repeated captions. Best viewed in colour.}
	\label{fig:diverse_examples}
\end{figure*}

In order to demonstrate contributions from different components in our model, we provide an extensive ablation study in Table~\ref{tab:control_ablation}.
We begin with baselines (Row 1 and 2) which are C-ST and C-BUTD model respectively.
Then in Row 3, we add the role-aware node embedding in the encoder and the performance is largely improved, which indicates that it is important to distinguish different intention roles in the graph.
Comparing Row 4 against Row 3 where the MR-GCN is employed for contextual graph encoding, we see that graph context is beneficial for the graph node encoding.
Row 5 and 6 enhance the decoder with graph flow attention and graph updating respectively.
The graph flow attention shows complementarity with the graph content attention via capturing the structure information in the graph, and outperforms Row 4 on two datasets.
However, the graph updating mechanism is more effective on MSCOCO dataset where the number of graph nodes are larger than on VisualGenome dataset.
Since the graph updating module explicitly records the status of graph nodes, the effectiveness might be more apparent when generating longer sentences for larger graphs.
In Row 7, we incorporate all the proposed components which obtains further gains.
Finally, we apply beam search on the proposed model and achieves the best performance.

Besides ASGs corresponding to groundtruth captions, in Figure~\ref{fig:user_intention_examples} we show an example of user created ASGs which represent different user intentions in a fine-grained level.
For example, ASG0 and ASG1 care about different level of details about the woman, while ASG2 and ASG5 intends to know relationships between various number of objects.
Subtle differences such as directions of edges also influence the captioning order as shown in ASG3 and ASG4.
Even for large complex graphs like ASG6, our model still successfully generates desired image captions.

\begin{table}
	\centering
	\small
	\caption{Comparison with state-of-the-art approaches for diverse image caption generation.}
	\label{tab:diversity_sota_cmpr}
	\begin{tabular}{c|l|ccc} \toprule
		& Method & Div-1 & Div-2 & SelfCIDEr \\ \midrule
		\multirow{2}{*}{\begin{tabular}[c]{@{}c@{}}Visual\\ Genome\end{tabular}} & Region & 0.41 & 0.43 & 0.47 \\
		& Ours & \textbf{0.54} & \textbf{0.63} & \textbf{0.75} \\ \midrule
		\multirow{5}{*}{\begin{tabular}[c]{@{}c@{}}MS\\COCO\end{tabular}} & BS \cite{aneja2019sequential} & 0.21 & 0.29 & - \\
		& POS \cite{Deshpande_2019_CVPR} & 0.24 & 0.35 & - \\
		& SeqCVAE \cite{aneja2019sequential} & 0.25 & 0.54 & - \\ \cmidrule{2-5}
		& BUTD-BS & 0.29 & 0.39 & 0.58 \\
		& Ours & \textbf{0.43} & \textbf{0.56} & \textbf{0.76} \\ \bottomrule
	\end{tabular}
	\vspace{-10pt}
\end{table}

\subsection{Evaluation on Diversity}
The bonus of our ASG-controlled image captioning is the ability to generate diverse image descriptions that capture different aspects of the image at different level of details given diverse ASGs.
We first automatically obtain a global ASG for the image (Section~\ref{sec:method_sgl}), and then sample subgraphs from the ASG.
For simplicity, we randomly select connected subject-relationship-object nodes as subgraph and randomly add one attribute node to subject and object nodes.
On VisualGenome dataset, we compare with dense image captioning approach which generates diverse captions to describe different image regions. 
For fair comparison, we employ the same regions as our sampled ASGs.
On MSCOCO dataset, since there are only global image descriptions for images,  we utilise beam search of BUTD model to produce diverse captions as baseline.
We also compare with other state-of-the-art methods \cite{aneja2019sequential,Deshpande_2019_CVPR} on MSCOCO dataset that strive for diversity.

As shown in Table~\ref{tab:diversity_sota_cmpr}, the generated captions of our approach are more diverse than compared methods especially on the SelfCider score \cite{wang2019describing} which focuses on semantic similarity.
We illustrate an example image with different ASGs in Figure~\ref{fig:diverse_examples}.
The generated caption effectively respects the given ASG, and the diversity of ASGs leads to significant diverse image descriptions.

\section{Conclusion}
In this work, we focus on controllable image caption generation which actively considers user intentions to generate desired image descriptions.
In order to provide a fine-grained control on what and how detailed to describe, we propose a novel control signal called Abstract Scene Graph (ASG), which is composed of three types of abstract nodes (object, attribute and relationship) grounded in the image without any semantic labels.
An ASG2Caption model is then proposed with a role-aware graph encoder and a language decoder specifically for graphs to follow structures of the ASG for caption generation.
Our model achieves state-of-the-art controllability conditioning on user desired ASGs on two datasets. It also significantly improves diversity of captions given automatically sampled ASGs.

{\small
\bibliographystyle{ieee_fullname}
\bibliography{reference}
}

\appendix



\section{Automatic ASG Generation}
Since the abstract scene graph does not require semantic labels, we could just utilize an off-the-shelf object proposal model to detect possible regions as object nodes.
The attribute and relationship nodes then can be added arbitrarily on or between object nodes because we can always describe attributes of an object or find certain relationship between two objects in the image.
However, not all relationships are meaningful and common to us. For example, ``a dog is chasing a rabbit'' is more common than ``a dog is chasing a computer''.
Therefore, we can optionally employ a simple relationship classifier to tell whether two objects contain a meaningful relationship.

We train the relationship classifier with annotations in groundtruth ASGs.
Instead of recognizing exact semantic labels which is rather challenging, we only predict three classes, with 0 for no relationship between two objects, 1 for subject-to-object relationship and 2 for object-to-subject relationship.
Three types of features are utilized for the prediction.
The first type is the global image appearance. The second type is the region visual features of the two objects respectively, and the third type is the feature for relative spatial location of the two objects.
We balance the ratio of different classes as 2:1:1 during training.

For inference, we firstly detect bounding boxes of objects and apply SoftNMS \cite{bodla2017soft} to reduce redundancy.
Then we utilize the trained relationship classifier for each pair of objects.
Two objects are considered to contain meaningful relationship if the probability of class 0 is below certain threshold (0.5 in our experiments) and the relationship of two objects are selected as class 1 or 2 according to the predicted probabilities.
In this way, we build a global ASG which contains abstract object and relationship nodes.

\begin{figure}
	\includegraphics[width=0.8\linewidth]{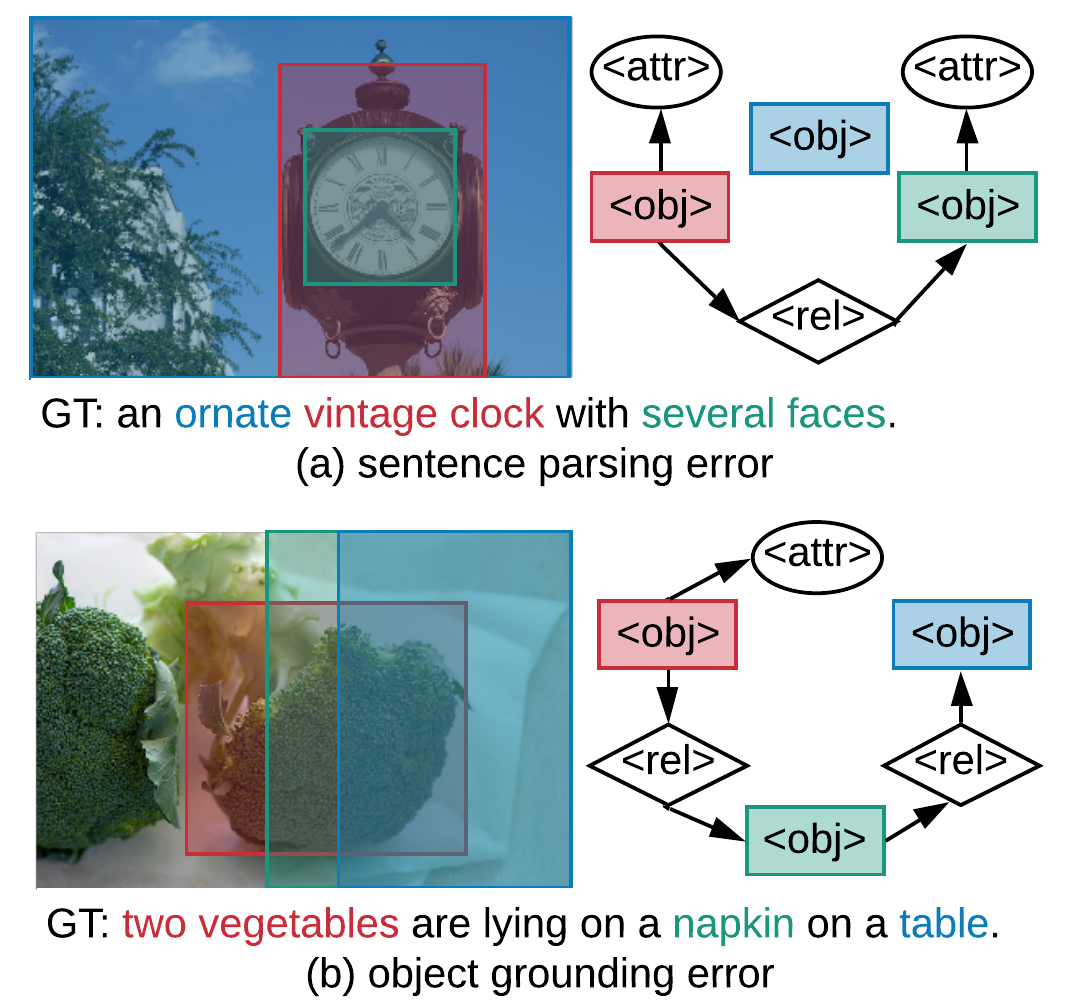}
	\caption{Two types of errors in the automatic dataset construction (examples from the testing set of MSCOCO).}
	\label{fig:annotation_error_examples}
\end{figure}

\section{ASG Dataset Construction}
For the VisualGenome dataset, although there are grounded region scene graphs for each region description, we notice that these region graphs are noisy with missing objects, relationships and misaligned attributes.
Therefore, we only utilize existing region scene graphs in VisualGenome as references to construct our ASGs. 
For the MSCOCO dataset, since there are no grounded scene graphs, we need to build  grounded ASGs from scratch.
The detailed steps of building an ASG $\mathcal{G}$ for image $\mathcal{I}$ and its image description $y$ are as follows:
\parskip=0.1em
\begin{enumerate}[itemsep=0.1em,parsep=0em,topsep=0em,partopsep=0em]
    \item utilize Stanford scene graph parser \cite{schuster2015generating} to parse description $y$ to a scene graph, where there are both semantic label and node type for each node and connections between nodes.
    \item collect candidate object bounding boxes and labels. For VisualGenome, we use the annotated object bounding boxes. For MSCOCO, we utilise an off-the-shelf object detector (Faster-RCNN pretrained on VisualGenome dataset) to detect objects.
    \item ground objects in the parsed scene graph to candidate object bounding boxes in the image. For VisualGenome, we take into account both location overlap between candidate objects and the region and semantic similarity of labels based on WordNet \cite{miller1995wordnet} for grounding. For MSCOCO, we can only utilize the semantic similarity of labels for grounding.
    \item remove noisy grounded scene graphs. If there are more than two objects in a scene graph without grounding, we remove the scene graph. For the remained scene graph, if an object cannot be grounded, we align the object with the region bounding box for VisualGenome and the global image for MSCOCO dataset.
    \item remove all semantic labels of nodes and only keep the graph layout and nodes type as our ASG $\mathcal{G}$.
\end{enumerate}

\begin{figure}
	\includegraphics[width=0.8\linewidth]{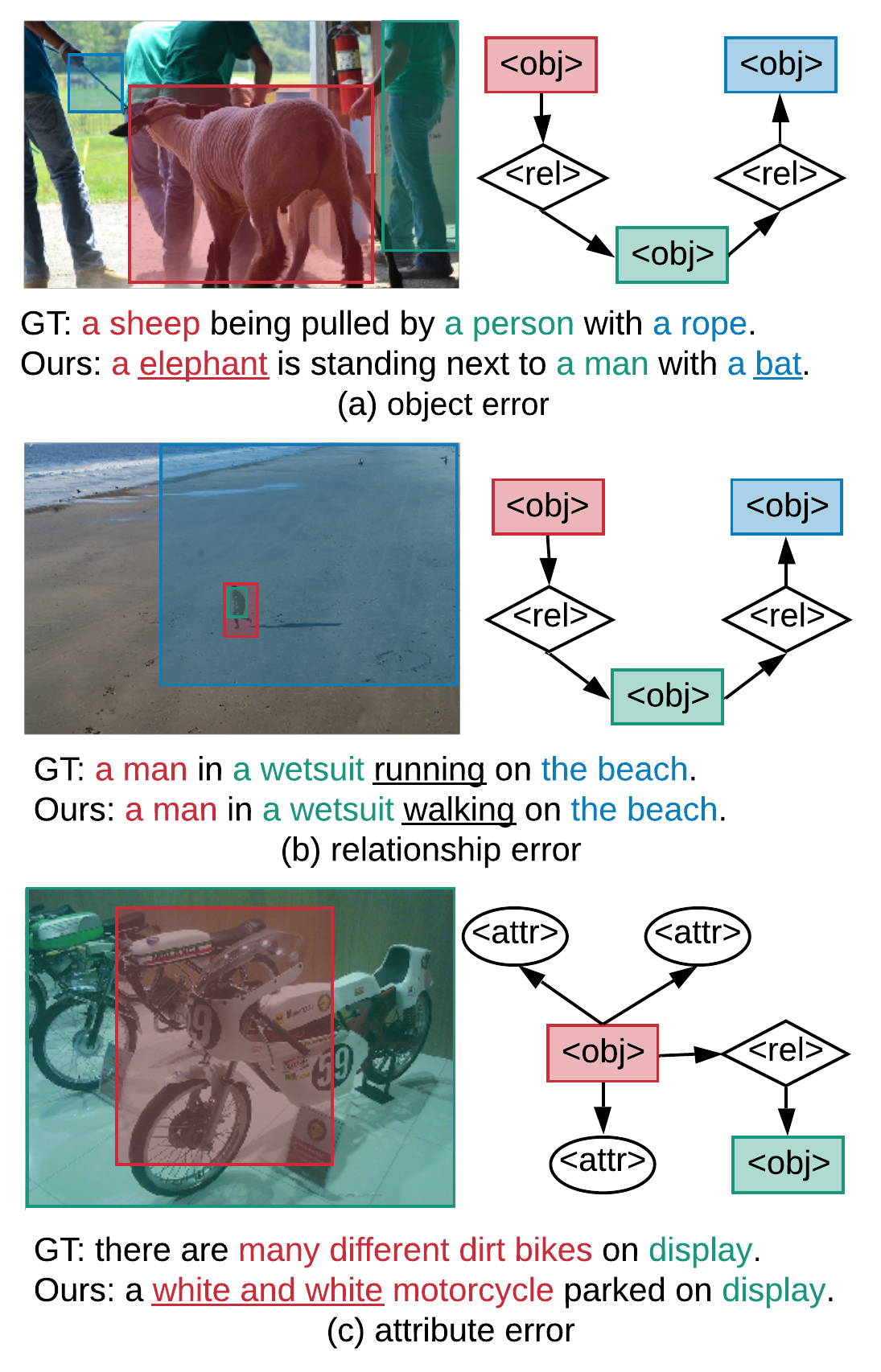}
	\caption{Three types of mistakes in our ASG2Caption model for controllable image caption generation (examples from the testing set of MSCOCO).}
	\label{fig:recognition_error_examples}
\end{figure}

To be noted, since the two datasets are automatically constructed, there mainly exists two types of noises especially for MSCOCO dataset where no object grounding annotations are available.
The two types of errors are sentence parsing error and object grounding error as shown in Figure~\ref{fig:annotation_error_examples}.
For example, in Figure~\ref{fig:recognition_error_examples} (a), the attribute ``ornate'' is mistaken as an object by incorrect sentence parsing; in Figure~\ref{fig:annotation_error_examples} (b), the object ``vegetables'' is only grounded on one broccoli but not two of them in the image.
However, since majority of the constructed pairs are correct, our model still can learn from the imperfect datasets. 

\section{Graph Structure Metric}
The proposed Graph Structure metric is based on SPICE metric \cite{anderson2016spice}.
The SPICE metric parses a sentence into three types of tuples $(o), (o, a)$ and $(o, r, o)$ and measures the semantic alignment of tuples between generated caption and groundtruth captions.
However, our Graph Structure metric only cares about the structure alignment which reflects the structure control of ASG without considering the semantic correctness.
For this purpose, we first calculate the numbers of the three types of tuples in the generated caption and groundtruth caption respectively.
Then we employ the mean absolute error for each tuple type as the structure misalignment measure, which is $G_o$, $G_a$, $G_r$ for measurement of $(o)$, $(o, a)$ and $(o, r, o)$ respectively.
The overall misalignment $G$ is the average of errors of the three tuple types.
The lower the score is, the better the structure alignment is.

\section{Additional Qualitative Results}
\begin{figure*}
	\includegraphics[width=1\linewidth]{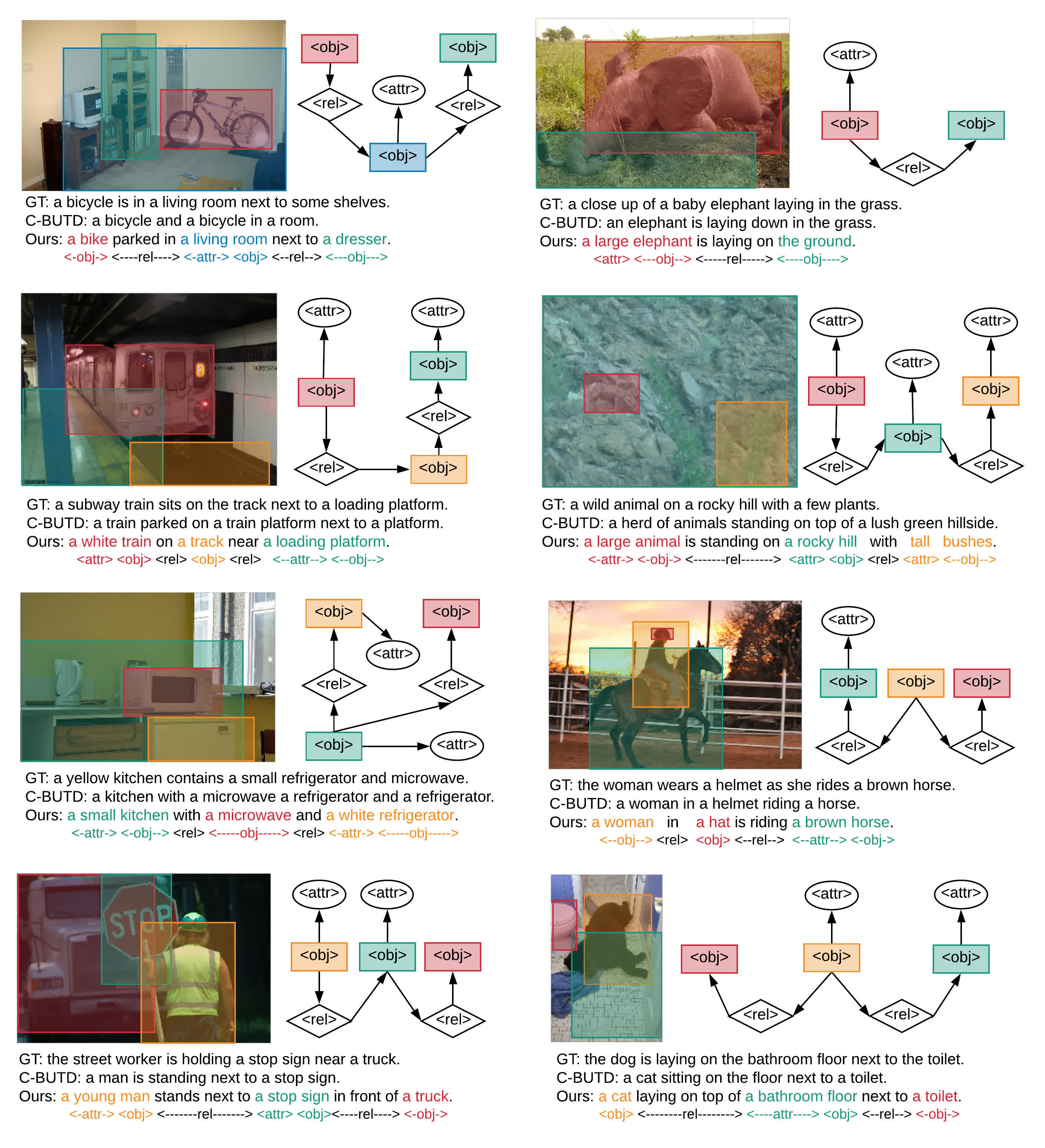}
	\caption{Examples for controllable image caption generation conditioning on designated ASGs compared with captions from the groundtruth and the state-of-the-art model C-BUTD \cite{anderson2018bottom}. Best viewed in color.}
	\label{fig:supp_control_examples}
\end{figure*}

\begin{figure*}
	\includegraphics[width=1\linewidth]{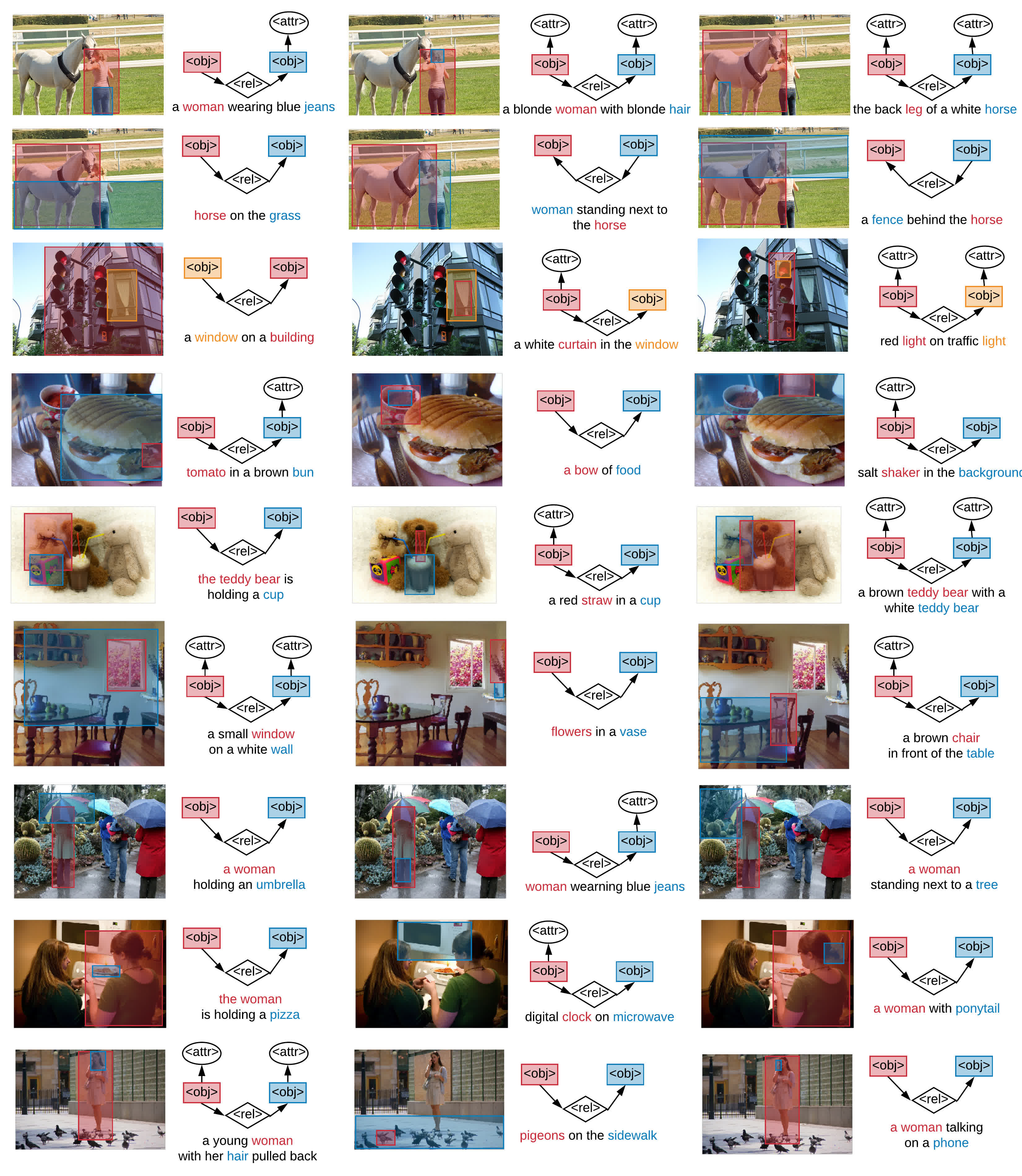}
	\caption{Examples for diverse image caption generation conditioning on sampled ASGs. Best viewed in color.}
	\label{fig:supp_diverse_examples}
\end{figure*}

Figure~\ref{fig:supp_control_examples} presents additional examples on controllable image caption generation with designated ASGs.
Figure~\ref{fig:supp_diverse_examples} provides more examples on diverse image caption generation with sampled ASGs.

In Figure~\ref{fig:recognition_error_examples}, we further present three main types of mistakes that our ASG2Caption model can make for controllable image caption generation, including object recognition error, relationship detection error and attribute generation error.
The attribute generation error mostly occurs when multiple attributes are required, which can lead to generation of repeated or incorrect attributes.

\end{document}